# Quantifying the Extent to Which Race and Gender Features Determine Identity in Commercial Face Recognition Algorithms

John J. Howard, Yevgeniy B. Sirotin, and
Jerry L. Tipton,
*The Maryland Test Facility*
`{john, yevgeniy, jerry}@mdtf.org`

Arun R. Vemury,
*The U.S. Department of Homeland Security,
Science and Technology Directorate*
`arun.vemury@hq.dhs.gov`

*Abstract*—Human face features can be used to determine individual identity as well as demographic information like gender and race. However, the extent to which black-box commercial face recognition algorithms (CFRAs) use gender and race features to determine identity is poorly understood despite increasing deployments by government and industry. In this study, we quantified the degree to which gender and race features influenced face recognition similarity scores between different people, i.e. non-mated scores. We ran this study using five different CFRAs and a sample of 333 diverse test subjects. As a control, we compared the behavior of these non-mated distributions to a commercial iris recognition algorithm (CIRA). Confirming prior work, all CFRAs produced higher similarity scores for people of the same gender and race, an effect known as "broad homogeneity". No such effect was observed for the CIRA. Next, we applied principal components analysis (PCA) to similarity score matrices. We show that some principal components (PCs) of CFRAs cluster people by gender and race, but the majority do not. Demographic clustering in the PCs accounted for only 10% of the total CFRA score variance. No clustering was observed for the CIRA. This demonstrates that, although CFRAs use some gender and race features to establish identity, most features utilized by current CFRAs are unrelated to gender and race, similar to the iris texture patterns utilized by the CIRA. Finally, reconstruction of similarity score matrices using only PCs that showed no demographic clustering reduced broad homogeneity effects, but also decreased the separation between mated and non-mated scores. This suggests it's possible for CFRAs to operate on features unrelated to gender and race, albeit with somewhat lower recognition accuracy, but that this is not the current commercial practice.

## I. Introduction

DURING the period from 2015 to 2020, face recognition (FR) experienced enormous increases in commercial investment, public interest, and public facing deployments. In 2014, convolutional neural nets applied to FR, achieved near human performance for the first time [1]. By 2016, at least two Fortune 500 companies began offering commercial facial recognition algorithms (CFRAs) via their cloud platforms [2] [3] and by the end of 2019, technology evaluations of face recognition algorithms by the U.S. National Institute of Standards and Technology (NIST) received 189 submissions from 99 distinct developers [4] (compared to 46 algorithms from 13 developers submitted for testing in their iris recognition (IR) program [5]. These commercial investments have spurred the deployment of numerous public facing services that leverage CFRAs for tasks such as boarding planes [6], finding suspected criminals [7], and buying beer [8]. There are numerous motivations for this growing public adoption, including the notion that human faces form a basis of identity for other humans and that humans perform face recognition on a daily basis, meaning most are familiar with the concept of FR.

However, recent reports have shown that CFRA performance can vary for people based on their demographic group membership [4] [9] [10]. One type of demographic variation observed is the tendency of CFRAs to assign greater similarity scores to different individuals that share gender and race categories. For example, comparing images of women to images of other women produces higher scores relative to scores produced when images of women are compared to images of men, an effect termed "broad homogeneity" [4] [9].

While intuitive based on human perception, this property of CFRAs can create undesirable behavior in many identification scenarios. For example, if an identification gallery, such as a most-wanted list, skews predominantly male, then men who are not in the gallery are more likely to be misidentified when searched against that gallery than women, solely on the basis of their male facial features.

In this study, we assess demographic variation in the performance of five CFRAs and one commercial iris recognition algorithm (CIRA). We first assess broad homogeneity effects, documented in [9], in our sample of

J. J. Howard, Y. B. Sirotin, and J. L. Tipton are with the Maryland Test Facility, Upper Marlboro MD 20774 USA (email: `{john, yevgeniy, jerry}` @ mdtf.org)

A R. Vemury is with the United States Department of Homeland Security, Science and Technology Directorate, Washington, DC, 20001 USA (email: arun.vemury@hq.dhs.gov).

algorithms, and then use a novel technique to quantify and compare these effects across black-box commercial algorithms.

## II. BACKGROUND AND SIGNIFICANCE

### A. Face Features

The human face has many features useful for identity. For example, intercanthal width is the distance between the inner portion of the eyes, and morphological nose width is the distance between the exterior nostrils [11]. The relative positions of *some* of these facial landmarks are shared by members of demographic groups. For example, the average male nose is shorter, broader, and more projecting relative to females [12] [13] and people of Sub-Saharan African ancestry tend to have broader noses than people of European and East Asian ancestry [14] [15]. However, other face features and their combinations are unlikely to be associated with gender or race. For instance, genetic disorders can be associated to specific common changes in face shape [16] [17] [18]. Likewise, features thought to be formed stochastically during development, such as iris texture utilized by iris recognition (IR) algorithms are unique not only to specific individuals, but to each eye [19]. Indeed, recent work indicates that gender and race features in face images can be manipulated while identity information relevant for face recognition is maintained [20] [21].

### B. The Consequences of Selecting Features Related to Protected Demographic Groups

All biometric samples inevitably share some common patterns. Biometric samples come from biological systems that may share some features due to common genetics, environment, or simply due to chance. When two biometric samples from different people are similar enough, biometric algorithms may label the two samples as matching, producing a false match. This can cause misidentifications, such as when the fingerprints from a 37 year old lawyer living in the U.S. state of Oregon, Brendan Mayfield, were incorrectly matched to a 32 year old Algerian man living in France, Daoud Ouhnane [22] [23].

Similarities in facial features are related to demographics, including gender and race (Section II.A). However, gender and race similarity *alone* are typically not enough to increase CFRA false match rates to unacceptable levels in most applications. Consequently, use of features related to gender and race has not been seen as a problem in the machine vision community. Nonetheless, small increases in one-to-one false match rate can lead to appreciable gains in one-to-N false positive identification rates, particularly when matching against large galleries [4]. This raises legitimate concerns about the fairness of CFRAs when matching against homogeneous galleries in law enforcement applications [24]. It is therefore important to understand the degree to which gender and race determine similarity scores produced by CFRAs.

### C. Evaluation of Commercial Algorithms versus Academic Algorithms

Generally, FR algorithms can be categorized as either commercial or academic. Much of the scientific literature, particularly around demographics in face recognition, focuses on academic algorithms [25] [26] [27]. The implementation details of academic algorithms are usually published and shared. However, leading commercial FR algorithms have superior performance relative to available academic algorithms [4] and come with the legal, financial, and operational support offered by commercial entities. Commercial face recognition algorithms are therefore used by industry and government to make real-world decisions [6] [7]. Consequently, the evaluation of commercial, not academic, algorithms should be paramount when discussing technology bias and fairness. However, unlike academic algorithms, commercial algorithms are "black-boxes", meaning little is known about the face template structure they produce or the inner workings of the FR algorithm. The only available information for evaluating commercial FR algorithm performance are the similarity scores they produce when comparing face images. This makes it necessary to develop methods of exploring demographic differentials that rely only on these similarity scores, and not training, template data, or mechanistic algorithm insight.

## III. METHODS

### A. Dataset

Data used in this study were collected during the 2018 DHS S&T Biometric Technology Rally [28]. Biometric samples were collected from 333 diverse test subjects on 11 different face and 5 iris biometric acquisition systems. All acquisition systems were commercially available systems from commercial biometric companies, available in 2018.

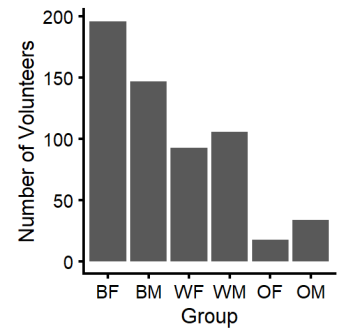

Fig. 1: : Number of volunteers by self-reported demographic race (B, Black or African American; W, White; O, all others) and gender (F, Female; M, Male).

The test described in [28] produced 3,324 face and 1,414 left iris probe images (all devices failed to acquire images on some subjects). For this study, these probe images were compared to galleries of 1,205 face images and 1,083 left iris images previously gathered from the same subjects over a six-year period from 2012-2018. Five different CFRAs and one CIRA were used to independently generate biometric similarity scores. In total, this operation produced 21,558,281 similarity scores which form the basis of this study. All matching systems were commercially available systems from established biometric companies, available in 2019. To comply with information sharing agreements between the test organizers and technology providers, all algorithm names are aliased in this report as "face1", "face2", "face3", "face4", "face5", and "iris".

Each algorithm produced an arbitrarily scaled similarity score for pairs of face or iris images. Larger scores corresponded to a greater likelihood that the two images belong to the same subject.

Demographic information, including race and gender, was self-reported by each of the 333 unique subjects (see Fig. 1). Most subjects in our sample self-identified as Black or African American, or White. For this reason, comparisons of same gender and race versus different race and gender groups was restricted to these demographic groups.

### B. Analysis Techniques

#### 1) 99th Percentile Non-Mated Score

Biometric false match rates are driven by the behavior of the tail of the non-mated distribution. We quantified the characteristics of this tail using shifts in the 99th percentile score of the imposter distribution, similar to [9]. In Equation 1, $S_{99,m}$ is the subject-specific 99th percentile non-mated score, $I_{(n)}(m)$ is the ordered set of non-mated similarity scores for subject $m$, and $n = \lceil .99 * |I| \rceil$.

$$S_{99,m} = I_{(n)}(m) \quad (1)$$

#### 2) Principal Component Analysis

The similarity scores described in Section III.A, were arranged into a matrix, per recognition algorithm, where each entry at row $i$ and column $j$ represented the average similarity score between subjects $i$ and $j$ for each of our 333 subjects. All score matrices were symmetric, with the diagonal of each matrix corresponding to the average mated score for each subject. To understand the individual variations with the strongest association to subject similarity, we performed principal components analysis (PCA) on the matrix produced by each algorithm. PCA is a linear dimensionality reduction technique. It can be used to transform high dimensional data into a series of principal components (PCs) in such a way that the highest level of variance is found on the first component, $PC_1$. Each subsequent $PC_k$ is orthogonal to the preceding and explains less variance ($\sigma_1^2 > \sigma_2^2 > \cdots > \sigma_k^2$). At some PC number $k$ the full or a sufficient amount of the cumulative variance ($\sum_k \sigma_k^2$) has been explained by the PCs. The remaining $n - k$ PCs can be discarded, thus accomplishing dimensionality reduction. In this study, PC decomposition of each score matrix and subsequent operations were performed using built-in functions available in the R statistical programming language [29].

PCA allows score matrices of different biometric algorithms to be compared using common units of explained variance. Recent prior work in demographics has measured performance variation in face recognition algorithms by comparing score distributions, error rates at fixed thresholds, and area under ROC curves (AUC) [4] [9] [30]. However, scores of different algorithms are scaled arbitrarily and error rates depend critically on thresholds, which must be determined separately for each algorithm. While AUC offers the ability to compare overall algorithm accuracy as a function of demographics, modern FR algorithms may make no errors on some datasets, producing uniform AUC (AUC = 1). Our measure allows algorithm comparisons in the absence of errors.

#### 3) Demographic Clustering

Each PC computed as described in Section III.B.2 corresponds to a pattern of score variation across 333 subjects. The similarity of face features between subjects in our dataset determines CFRA similarity scores. The PCs that explain the most variance for each algorithm correspond to the shared feature patterns that are most heavily weighted by each algorithm in determining similarity. We assessed the degree of association of these features with gender and race by measuring the distribution of these groups across each PC. Specifically, we measured the degree of demographic clustering by calculating a clustering index $C_k$ for each $PC_k$ by taking the ratio of within group deviation from the mean to overall deviation from the mean across all subjects $i$ in our sample according to Equation 2, where $D$ is the set of subjects belonging to a specific demographic group and $x_i$ is the value for subject $i$ on the PC.

$$C_k = \frac{\sum_D \sum_{i \in D}(x_i - \bar{x}_D)^2}{\sum_i (x_i - \bar{x})^2} \quad (2)$$

We assessed whether the clustering index value for each $PC_k$ was statistically significant by comparing the calculated $C_k$ values, which rely on the variance between subjects in real demographic groups ($\bar{\sigma}_{D,k}^2 = \frac{1}{N} \sum_D \sum_{i \in D}(x_i - \bar{x}_D)^2$), to the 99th percentile of the distribution of $C_{null}$, where $C_{null}$ is calculated by 500 shuffles assigning subjects to randomized demographic groups $D$. Given 333 PCs with no significant clustering, this criterion would, by chance label 3 as clustered.

Finally, to assess the overall demographic clustering for an algorithm, we measured the proportion of total variance in scores explained by demographic clustering according to Equation 3 where $\sigma_k^2$ is the variance of $PC_k$, $\sigma_{tot}^2$ is the total variance across the entire dataset, and $C_k$ is as described in Equation 2.

$$C_{tot} = \frac{1}{\sigma_{tot}^2} \sum_k \sigma_k^2 C_k \quad (3)$$

#### 4) D Prime Analysis

Since the PCs of algorithm similarity score matrices are orthogonal, it's possible to discard certain PCs and reconstruct score matrices as if these components did not exist. The reconstructed score matrices will have different distributions of mated (diagonal) and non-mated similarity (non-diagonal) scores. To quantify the separation between these two distributions, and the impact of this reconstruction step, we use the d-prime metric [31]. Previous studies of demographic effects in face recognition have also measured broad relative shifts in mated and non-mated distributions using d-prime [27]. We calculated the d-prime according to Equation 4 where $\mu$ and $\sigma^2$ are the mean and variance, and $M$ and $NM$ refer to the mated and the non-mated distributions of average similarity scores, respectively.

$$d' = \frac{\mu_M - \mu_{NM}}{0.5\sqrt{\sigma_M^2 - \sigma_{NM}^2}} \quad (4)$$

## IV. RESULTS

### A. Consistent Effects of Broad Demographic Homogeneity across Commercial Face Recognition Algorithms

Prior work has shown, using a single CFRA, that the tail of the non-mated similarity score distribution between subjects of the same gender and race is higher than the tail of the distributions between subjects of different genders and race [9]. All five CFRAs in our sample reliably followed this broad homogeneity effect (Fig. 2) Conversely, no effect of gallery homogeneity was observed for the CIRA.

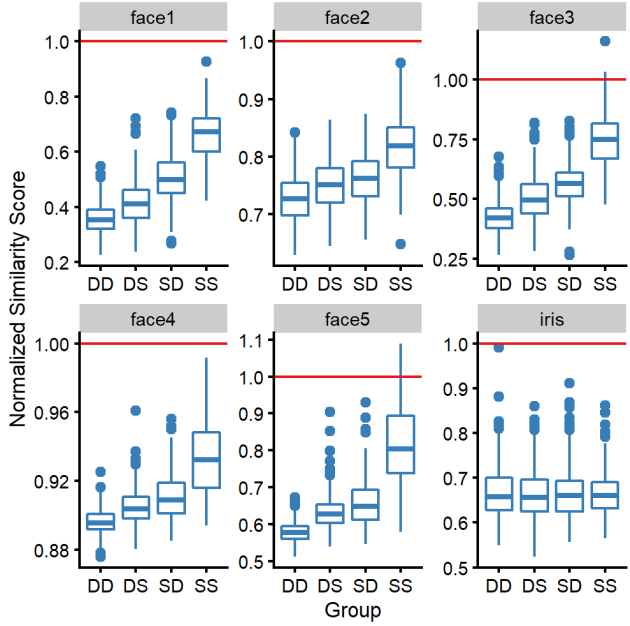

Fig. 2: Group homogeneity strongly modulates the tail of the non-mated distribution. Each facet corresponds to a different biometric algorithm and plots the 99th percentile of the non-mated score distribution (Section III.B.1) across individuals. Scores on the y-axis for each algorithm are divisively normalized such that scores at the 1:10,000 threshold (red line) get a value of 1. Groups along the x-axis are as follows: DD, different gender and race; DS, different gender and same race; SD, same gender and different race; SS, same gender and race.

### B. Face Recognition Score Matrices have Block-Diagonal Demographic Structure

Faces of different pairs of subjects have different features in common, only some of which are relevant to face recognition algorithms (Section II.A). The patterns of similarity scores for individuals known to share various features can reveal how these features are weighted by the algorithm in calculating face similarity. Variation in CFRA similarity scores is driven both by face features as well as by the properties of the images used in the comparison [10]. To isolate the effect of face features for each algorithm, we computed 110,889 average similarity scores between each unique pair of the 333 subjects in our sample.

Fig. 3 plots these average subject-to-subject similarity scores as a score matrix with rows and columns sorted based on the gender and race of each subject in our dataset (Section III.B.2).

Each score in this matrix is an average of 72 similarity scores between probe and gallery face images of the subjects and 28 similarity scores between probe and gallery left iris images of the subjects. As expected from Fig. 2, CFRA score matrices showed a clear block-diagonal structure with higher similarity scores for subject pairs within the same demographic group than between subjects in different demographic groups. This structure indicates the presence of correlations in the data that could be leveraged by a dimensionality reduction technique, such as PCA.

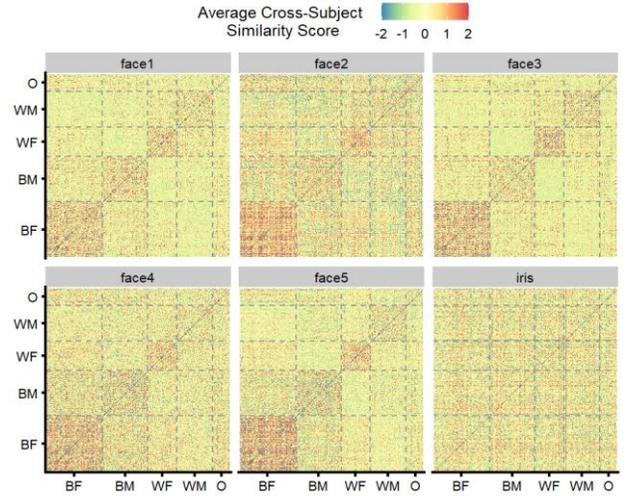

Fig. 3: CFRAs produce higher similarity scores within demographic groups. Each facet shows a raster plot of the average similarity scores produced for each pair of individuals in our sample. To aid visualization, scores have been normalized such that the non-mated scores have $\mu = 0$ and $\sigma = 0$. Dashed lines separate demographic groups. Note block diagonal structure present for all CFRAs, but not for the CIRA

### C. Face Recognition Algorithms Cluster Individuals by Race and Gender

Fig. 2 and Fig. 3 suggest that all CFRAs show homogeneity effects. However, it is difficult to compare the magnitude of the effects across algorithms because similarity scores returned by each black-box CFRA are scaled arbitrarily. We used PCA (Section III.B.2) to reduce the dimensionality of the similarity metric (Fig. 4) and isolate the contribution of particular components to the overall variation in the data.

After applying PCA, each PC corresponds to a score pattern across individuals in our sample. Assuming that score patterns are related to the face features of individuals, those patterns that explain the largest proportion of similarity score variance should therefore separate subjects based on the relative contribution of this feature to score variance. For instance, if similarity scores were determined solely by the relative width of the nose, then our subjects would be ordered based on nose width along the first PC of the score matrix. If, on the other hand, scores were not related to nose length, but rather related to the intercanthal width, then subjects would be ordered by distance between the eyes and not by nose length. Though we cannot know the important features used by the black-box CFRAs, we can examine the extent to which the order of subjects along each PC corresponds to demographic groups.

Further, since each PC has a known contribution to overall score variance, we can quantify the extent to which known demographic categories determine the similarity scores.

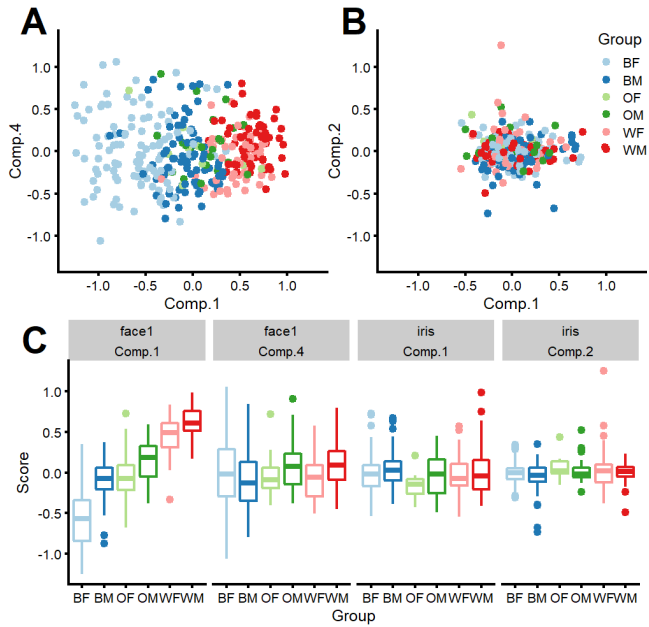

Fig. 4: Visualization of select principal components. **A.** Component 1 for algorithm face1 shows distinct clustering by demographic group, but component 4 does not. **B.** Components 1 and 2 for algorithm iris do not show demographic clustering. **C.** Distributions of component values visualized associated with different demographic groups.

### D. Comparing Demographic Clustering Across Commercial Face Recognition Algorithms.

We quantified the clustering illustrated in Fig. 4 by computing a clustering index for each PC (Section III.B.3, Equation 2). The clustering index is bounded between 0 and 1, with zero signaling that the variance within each gender and race group is the same as overall variance. A clustering index of 1 indicates that there is no variance across individuals within each gender and race group.

Fig. 5 shows the clustering index for the first ten PCs of each algorithm. All five CFRAs showed statistically significant clustering for the first two PCs according to the test described in Section III.B.3. Additionally, the first two PCs explained between 12 and 27 percent of the variance in similarity scores, depending on CFRA. None of the first ten PCs had significant demographic clustering for CIRA similarity scores.

To compare the extent to which different algorithms exhibited demographic clustering, we next measured the clustering index across all PCs (Section III.B.3, Equation 3). On average, we found demographic clustering accounted for 10% of total CFRA score variance, ranging from 6% for "face4" to 14% for "face3" (Fig. 5B). Clustering accounted for less than 2% of the variance in similarity scores produced by the CIRA. Of the 333 PCs calculated for the CFRAs, on average 14 showed significant clustering, compared with one for the CIRA (Fig. 5C). Components with no significant clustering accounted, on average for 62% of total score variance for CFRAs. These components reflect face feature variances that is not associated with gender or race.

### E. Estimating the Effects of Ignoring Demographically Clustered Features

We estimated the potential performance impact of having CFRAs ignore face features associated with gender and race. To do this, we reconstructed average similarity score matrices after removing all components with significant clustering and then compared the effects on the mated and non-mated distributions using the $d'$ statistic (Section III.B.4). Fig. 6 shows the distributions of average similarity scores in the original and reconstructed score matrices. As expected, removing PCs with significant clustering brought the non-mated distributions of average scores between subjects of the same gender and race (SS) closer to the non-mated average scores between subjects of different genders and races (DD). However, the operation also brought the overall mated and non-mated distributions closer together, decreasing $d'$. Nonetheless, even after reconstruction, $d'$ values remained high for some algorithms.

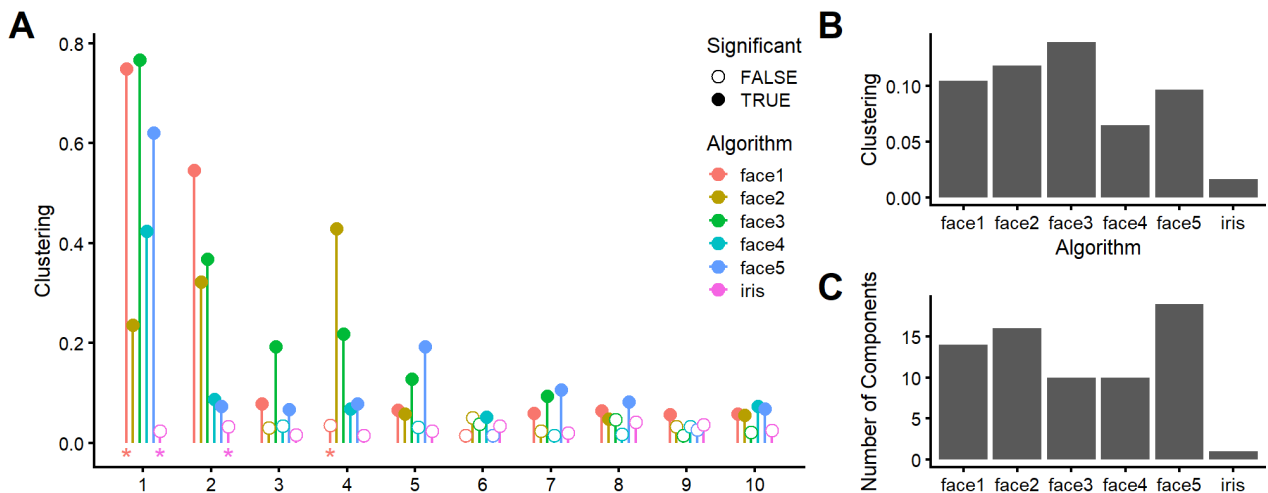

Fig. 5: Quantification of demographic clustering across algorithms. **A.** Stem plot of the demographic clustering index computed for each component. Filled circles correspond to components with statistically significant clustering (Section III.B.3). Asterisks mark components visualized in Fig. 4. **B.** The total proportion of similarity score variance explained by demographic clustering for each algorithm. **C.** Number of principal components with statistically significant clustering for each algorithm $\alpha = 0.01$.

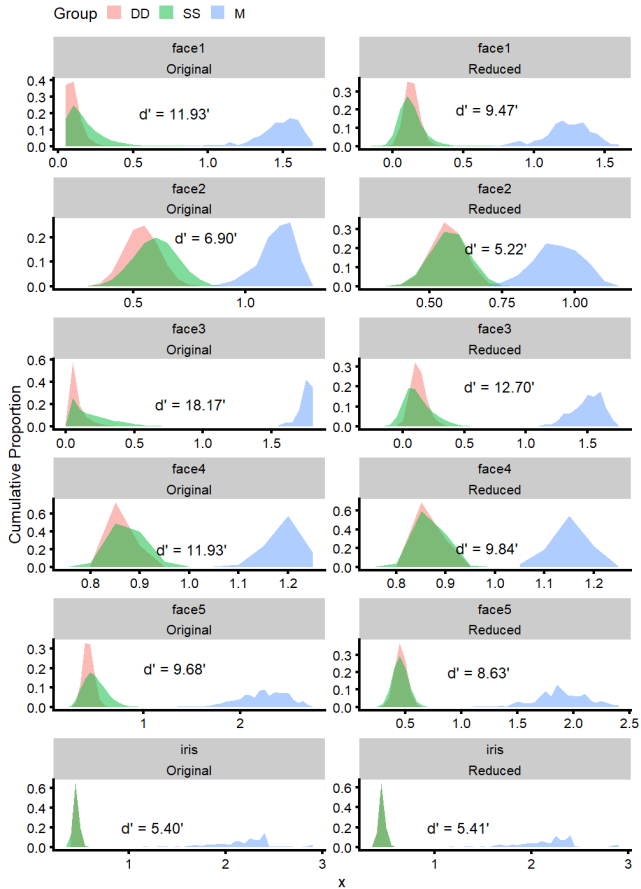

Fig. 6: Performance impact of removing components with significant demographic clustering. Each facet shows the empirical distributions of mated and non-mated scores in the original score matrix and after dimensionality reduction. Distributions are color coded as: SS, same gender and race; DD, different gender and race; M, mated. The SS and DD non-mated distributions include only individuals identifying as "Black" or as "White". Values of d' are based on the comparison of the mated distribution (M) to the full non-mated distribution (SS+DD). Note that SS and DD distributions for reduced scores are closer together. Note also that M and the overall non-mated distribution move closer in the reduced face recognition matrices as quantified by $d'$.

This suggests that ignoring face features associated with gender and race is likely to only modestly reduce CFRA performance since the range of original CFRA $d'$ values (6.90 to 18.17) largely overlap the range of reduced $d'$ values (5.22 to 2.70).

## V. DISCUSSION

In this paper, we discuss the extent to which five commercial face recognition algorithms (CFRAs) and one commercial iris recognition algorithm (CIRA) utilize face features associated with gender and race in determining individual identity. We first show that, non-mated similarity scores of all five CFRAs were higher between subjects of the same gender and race.

Quantifying the proportion of score variance explained by gender and race information across CFRAs using principal component analysis, we show that some principal components cluster individuals by race and gender whereas most do not.

Use of face features associated with gender and race by CFRAs creates concerns regarding the fairness of these algorithms in some applications. Recent work by privacy groups [24] highlighted the fact that law enforcement face image galleries can be demographically homogeneous, with African American Males comprising a majority of the faces. The demographic clustering documented in this research means that performing identifications against such galleries using images of out-of-gallery African-American Males would yield higher rank-1 similarity scores relative to White Females. To the extent that our sample of CFRAs is representative, it suggests that use of the current generation of CFRAs to perform identifications against large homogeneous galleries can result in disparate treatment based on race and gender [32].

However, our research also shows that this outcome is likely avoidable. We found that most variation in CFRA similarity scores is not associated with race and gender. Further, separation between mated and non-mated score distributions reconstructed exclusively using PCs that do not cluster individuals by race and gender was only modestly reduced, suggesting CFRAs can maintain acceptable performance even when ignoring face features associated with race and gender. Indeed, recent work suggests that demographic features can be removed from face images while maintaining subsequent face recognition [20] [21]. This is what has long been observed in iris recognition. The periocular images used in iris recognition bear features related to demographics and both humans and algorithms can readily identify race and gender from periocular images [33] [34] [35]. Nonetheless, iris recognition algorithms based on iris-codes do not utilize these features in making identity determinations [19].

Our research suggests caution when using current CFRAs when performing identifications against large, homogeneous galleries and points to a need for audits of operational systems to measure the extent to which the differential performance demonstrated here leads to differential outcome in operational use. Human review with orthogonal information may mitigate such occurrences. Developing demographically-blind CFRAs that explicitly ignore face features associated with race and gender will help maintain fairness as use of this technology grows. We believe that developing such algorithms and demonstrating fairness, including reduced demographic clustering, should be a focus for companies selling face recognition technology.

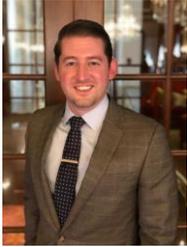

**Dr. John J. Howard** (M'09) received his Ph.D. in Computer Science from Southern Methodist University. His thesis was on pattern recognition models for identifying subject specific match probability. His current research interests include biometrics, computer vision, machine learning, testing human machine interfaces, pattern recognition, and statistics. He has served as the principal investigator on numerous R&D efforts across the intelligence community, Department of Defense, and other United States Government agencies. He is a member of the SAIC Identity and Data Sciences Lab and currently the Principal Data Scientist at the Maryland Test Facility.

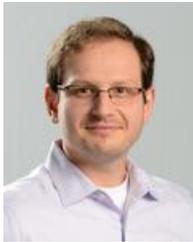

**Dr. Yevgeniy B. Sirotin** holds a Ph.D. in Neurobiology and Behavior from Columbia University and has diverse research interests in behavior and human computer interaction. His past research spans mathematical psychology (cognitive modeling), neurophysiology (multi-spectral imaging of the brain), psychometrics (mechanisms of visual and olfactory perception), biometrics (design and testing of identity systems), and human factors (usability). He currently works as Principal Investigator and Manager of the Identity and Data Sciences Laboratory at SAIC which supports applied research in biometric identity technologies at the Maryland Test Facility.

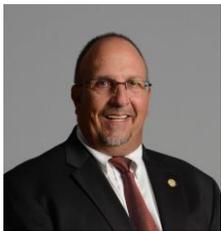

**Jerry L. Tipton.** is the Program Manager and Director of SAIC's Identity and Data Sciences Lab. He has over 20 years' experience in the biometric industry with over 15 years managing research portfolios in support of various United States Government agencies. He currently supports the Department of Homeland Security, Science and Technology Directorate at the Maryland Test Facility.

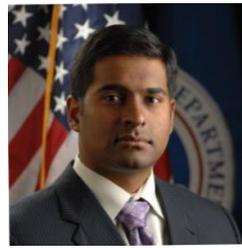

**Arun R. Vemury.** received his M.S. in Computer Engineering from George Washington University. His current research interests include biometrics, pattern recognition, machine learning, and operations research. He serves as the Director of the Biometrics and Identity Technology Center for the United States Department of Homeland Security Science and Technology Directorate.